\documentclass[sigconf, screen, nonacm]{acmart}

\definecolor{lightblue}{rgb}{0.68, 0.85, 0.9}
\definecolor{lightcyan}{rgb}{0.88, 1.0, 1.0}

\newcommand{\mff}[1]{\underline{\textbf{#1}}}

\newcommand{\muu}[1]{\textcolor[HTML]{2D8000}{#1}}
\newcommand{\mdd}[1]{\textcolor[HTML]{D90429}{#1}}

\newcommand{\model}{\textsl{SynC}}
\newcommand{\plusmodel}{+\model}

\newcommand{\Tref}[1]{Table~\ref{#1}}
\newcommand{\Eref}[1]{Eq.~(\ref{#1})}
\newcommand{\Fref}[1]{Fig.~\ref{#1}}

\newcommand{\Sref}[1]{Sec.~\ref{#1}}

\newcommand{\sonetoone}{S_\text{One}}
\newcommand{\sonetomany}{S_\text{T2I}}

\newcommand{\s}{S}
\newcommand{\dsyn}{\mathcal{D}_{syn}}
\newcommand{\Baseline}{\textit{Baseline}}

\newcommand{\fclip}{f_\text{CLIP}}

\newcommand{\fret}{f_\text{ret}}
\newcommand{\fsigliptwo}{f_\text{SigLIP2}}
\newcommand{\f}{f}

\newcommand{\onevar}{(\cdot)}
\newcommand{\twovar}{(\cdot,\cdot)}

\newcommand{\rttot}{R}
\newcommand{\ritot}{\hat{R}}

\newcommand{\ven}[1]{$_{\demph{\text{#1}}}$}
\definecolor{mainlinecolor}{HTML}{E8F3EE}
\newcommand{\lineColor}{mainlinecolor}
\definecolor{appendixlinecolor}{RGB}{230,230,230}

\definecolor{demphcolor}{RGB}{144,144,144}
\newcommand{\demph}[1]{\textcolor{demphcolor}{#1}}

\definecolor{variantcolor}{HTML}{75AB75}
\newcommand{\VariantColor}{variantcolor}

\usepackage{multirow}
\usepackage{xcolor,colortbl}
\usepackage{makecell}

\AtBeginDocument{%
  }

\setcopyright{acmlicensed}
\copyrightyear{2025}
\acmYear{2025}
\acmDOI{XXXXXXX.XXXXXXX}
\acmConference[MM'25]{the 33rd ACM International Conference on Multimedia}{Oct 27--31,
  2025}{Dublin, Ireland}
\acmISBN{978-1-4503-XXXX-X/YY/MM}

\begin{document}

\title{SynC: Synthetic Image Caption Dataset Refinement with One-to-many Mapping for Zero-shot Image Captioning}


\author{Si-Woo Kim}
\affiliation{%
  \institution{Hanyang University}
  \city{}
  \country{}
  }
\email{boreng0817@hanyang.ac.kr}

\author{MinJu Jeon}
\affiliation{%
  \institution{Hanyang University}
    \city{}
  \country{}
  }
\email{mnju5026@hanyang.ac.kr}

\author{Ye-Chan Kim}
\affiliation{%
  \institution{Hanyang University}
    \city{}
  \country{}
  }
\email{dpcksdl78@hanyang.ac.kr}

\author{Soeun Lee}
\affiliation{%
  \institution{AI R\&D Division, CJ Group}
    \city{}
  \country{}
  }
\email{dlth508@cj.net}

\author{Taewhan Kim}
\affiliation{%
  \institution{Hanyang University}
    \city{}
  \country{}
  }
\email{taewhan@hanyang.ac.kr}

\author{Dong-Jin Kim}
\authornote{Corresponding author.}
\affiliation{%
  \institution{Hanyang University}
    \city{}
  \country{}
  }
\email{djdkim@hanyang.ac.kr}

\newcommand{\mjjeon}[1]{\textcolor[rgb]{1,0,0}{#1}}

\renewcommand{\shortauthors}{Kim et al.}
%\renewcommand{\shortauthors}{Si-Woo Kim, MinJu Jeon, Ye-Chan Kim, Soeun Lee, Taewhan Kim, \& Dong-Jin Kim}

      




\begin{abstract}
Zero-shot Image Captioning (ZIC) increasingly utilizes synthetic datasets generated by text-to-image (T2I) models to mitigate the need for costly manual annotation. However, these T2I models often produce images that exhibit semantic misalignments with their corresponding input captions (e.g., missing objects, incorrect attributes), resulting in noisy synthetic image-caption pairs that can hinder model training. Existing dataset pruning techniques are largely designed for removing noisy text in web-crawled data. However, these methods are ill-suited for the distinct challenges of synthetic data, where captions are typically well-formed, but images may be inaccurate representations.
To address this gap, we introduce $\model$, a novel framework specifically designed to refine synthetic image-caption datasets for ZIC. Instead of conventional filtering or regeneration, $\model$ focuses on reassigning captions to the most semantically aligned images already present within the synthetic image pool. Our approach employs a one-to-many mapping strategy by initially retrieving multiple relevant candidate images for each caption. We then apply a cycle-consistency-inspired alignment scorer that selects the best image by verifying its ability to retrieve the original caption via image-to-text retrieval.
Extensive evaluations demonstrate that $\model$ consistently and significantly improves performance across various ZIC models on standard benchmarks (MS-COCO, Flickr30k, NoCaps), achieving state-of-the-art results in several scenarios. $\model$ offers an effective strategy for curating refined synthetic data to enhance ZIC.\footnote{Code: \url{https://github.com/boreng0817/SynC}}
\end{abstract}    

\maketitle

\section{Introduction}

High-quality image-caption paired datasets~\cite{coco} are crucial for training effective image captioning models. However, creating such datasets at scale through manual annotation is resource-intensive and laborious. Consequently, Zero-Shot Image Captioning (ZIC) has emerged as a prominent research direction. Initial ZIC approaches~\cite{decap, capdec, viecap, ifcap} primarily leveraged the cross-modal understanding capabilities of models like CLIP~\cite{clip} for text-only training. More recently, the advent of powerful text-to-image (T2I) generative models, capable of synthesizing photorealistic images~\cite{ldm, ramesh2021zero}, has spurred a new wave of ZIC methods~\cite{pcm-net, syntic, icsd}. These approaches generate synthetic images from text corpora using models like Stable Diffusion (SD) \cite{ldm}, thereby creating image-caption pairs for training captioning models without manual annotation.

\begin{figure*}[t]
    \centering    
    \includegraphics[width=\linewidth]{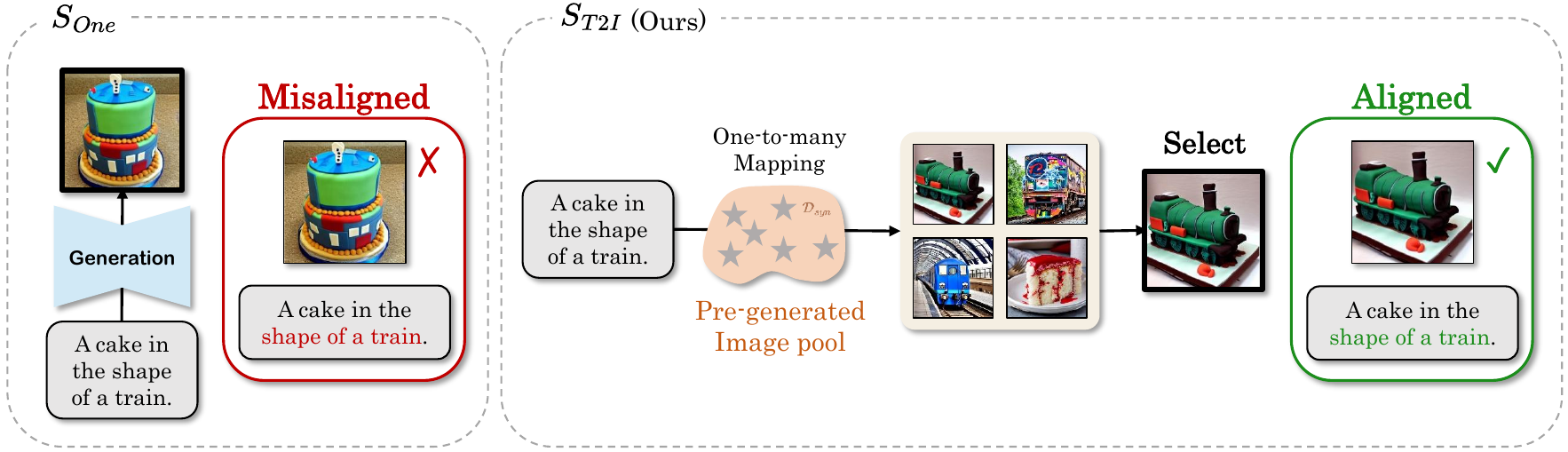}
    \caption{Previous researches~\cite{pcm-net, synthcap, syntic} employ $\sonetoone\onevar$ to construct synthetic image caption pairs for training. The text-to-image generative model may fail to synthesize the desired details of the query caption. Our proposed one-to-many mapping strategy $\sonetomany\onevar$ selects multiple relevant candidates for the given query caption. Utilizing an accurate multi-modal scorer function with $\sonetomany\onevar$ can refine misaligned pairs, as shown in the right part of the figure.}
    \label{fig:teaser_one_vs_t2t}
\end{figure*}

%Despite the promise of T2I models, reliably generating images that accurately reflect the semantics of complex input captions remains a significant challenge~\cite{lee2024holistic, conwell2022testing, bakr2023hrs}. 
Despite the advances of T2I models, generating images that accurately reflect the semantics of complex input captions remains a significant challenge~\cite{lee2024holistic, conwell2022testing, bakr2023hrs}. As illustrated in the left part of~\Fref{fig:teaser_one_vs_t2t}, T2I models like SD often produce images exhibiting semantic misalignment with the source prompt, such as omitting specified objects or generating incorrect attributes. Existing T2I-based ZIC methods~\cite{pcm-net, icsd} attempt to mitigate the issue of poorly generated objects during training~\cite{pcm-net} or optimize prompt engineering by summarization leveraging Large Language Models (LLMs)~\cite{icsd}. However, these strategies do not explicitly guarantee or verify the semantic alignment between the final synthetic image and its corresponding caption. As a result, the synthetic datasets used by these methods~\cite{pcm-net, icsd} inevitably contain misaligned image-caption pairs, potentially hindering model training.

This necessitates effective strategies for curating synthetic image-caption datasets. However, existing dataset filtering techniques are ill-suited for this specific challenge. Prior work on filtering captioning datasets has primarily focused on different objectives, such as gender debiasing in real-world datasets~\cite{hirota2023model} or curating paired data from unpaired sources~\cite{SS1M, kim2019image, kim2024semi}. Furthermore, techniques developed for pruning noisy web-scraped Vision-Language Model (VLM) datasets~\cite{sieve, laclip, veclip, clips, synthclip} are also not directly applicable. This is because existing methods typically address noise originating from low-quality alt-text paired with real-world images~\cite{improving, sharma2018conceptual}. In contrast, synthetic datasets exhibit the inverse characteristic: the captions are generally well-formed, but the images are the source of noise due to generation inaccuracies. Applying these VLM pruning methods~\cite{veclip, laclip, synthclip} can even degrade ZIC performance and incur significant computational overhead, partly because several methods~\cite{veclip, synthclip, sieve} rely on powerful pretrained captioning models (e.g., BLIP~\cite{li2022blip}, LLaVA~\cite{llava}), raising fairness issues or LLMs~\cite{laclip} which cannot inherently fix image-based misalignments.

To address the unique challenges of synthetic dataset curation for ZIC, we propose \textbf{$\model$}, a novel framework for refining synthetic image-caption pairs. Instead of iteratively regenerating images until alignment is achieved~\cite{sariyildiz2023fake}, $\model$ focuses on reassigning captions to existing synthetic images to maximize semantic alignment. Our approach involves two key components: (1) A \textit{one-to-many mapping strategy} based on a text-to-image (T2I) retrieval mechanism with a caption as a query, where each caption retrieves multiple potentially relevant synthetic images using T2I retrieval. (2) A  \textit{multi-modal alignment scoring function} to select the most relevant synthetic image among candidate images selected with the proposed one-to-many mapping strategy. Inspired by cycle consistency learning frameworks~\cite{zhao2025doracycle, li2023leveraging}, this scorer evaluates alignment by checking if a candidate image can reliably retrieve its corresponding caption (or semantically similar captions) via image-to-text (I2T) retrieval. With the combination of the T2I retrieval-based one-to-many mapping strategy and the I2T retrieval-based multi-modal alignment scoring function, $\model$ identifies a highly relevant synthetic image for the query caption via dual cross-modal retrieval, T2I for candidates, and I2T for scoring. $\model$ refines noisy synthetic image-paired datasets by retaining newly assigned synthetic image pairs with high alignment scores, which further improves the performance of zero-shot image captioning models.

Despite its conceptual simplicity, $\model$ demonstrates consistent and significant performance improvements across various ZIC models~\cite{pcm-net, ifcap, viecap, capdec} in standard image captioning benchmarks, including MS-COCO, Flickr30k, and NoCaps~\cite{coco, flickr30k, nocaps}. Our extensive experiments validate the effectiveness of $\model$ for refining synthetic image-caption datasets, particularly for enhancing ZIC model training. Furthermore, $\model$ provides practical guidelines for leveraging large-scale synthetic datasets generated by T2I models in the context of zero-shot image captioning.

In summary, our contributions are as follows:
\begin{itemize}
\item We address the quality and curation challenges of synthetic images for zero-shot image captioning by proposing a novel refinement framework, \textbf{$\model$}, which employs a flexible, cycle-consistency inspired approach to refine more accurate image-text pairs.
\item We design \textit{one-to-many mapping strategy} and \textit{alignment score function} for synthetic image selection to evaluate image-caption alignment, enabling accurate image-caption pairing through selecting semantically aligned images and filtering noisy images.
\item Extensive evaluations demonstrate that $\model$ achieves consistent performance improvements in various zero-shot image captioning models, even reaching state-of-the-art performance in various scenarios.
\end{itemize}\label{sec:intro}

\section{Related Work}

\noindent\textbf{Zero-shot Image Captioning.} Recent Zero-Shot Image Captioning (ZIC) methods increasingly utilize Text-to-Image (T2I) models like Stable Diffusion~\cite{ldm} to synthesize training data~\cite{pcm-net, icsd, syntic}, bypassing the need for manual annotations. For instance, ICSD~\cite{icsd} employs Large Language Models (LLMs) for prompt summarization and selection before generating images, creating a synthetic dataset analogous to standard benchmarks~\cite{coco, flickr30k}. PCM-Net~\cite{pcm-net} also generates synthetic datasets but uses captions directly as prompts and incorporates training strategies to mitigate the quality issues inherent in synthetic images. While previous synthetic image-based ZIC methods address the quality of generated images, they often overlook subtle misalignments between generated images and their corresponding captions, particularly in capturing fine-grained details. In contrast, our $\model$ incorporates a flexible one-to-many mapping strategy and cycle-consistency inspired alignment score, enabling more accurate image-text pairing and significantly improving zero-shot captioning performance.

\noindent\textbf{Dataset Pruning.} Filtering web-crawled dataset is crucial for training high-performance Vision-Language Models (VLMs)~\cite{improving, sieve, laclip, veclip, synthclip, wang2024variance, clips}. Existing pruning methods typically modify text (augment-based~\cite{improving, laclip, veclip, synthclip}) or filter pairs based on initial similarity scores (metric-based~\cite{sieve, clips, wang2024variance}); these approaches focus primarily on the textual modality, as web-crawled captions and alt-text are often noisy. This focus has proven effective for web data and enhances VLM performance. However, synthetic image-caption datasets possess distinct characteristics. The captions are typically well-formed, but the challenge lies in the potential visual-semantic mismatch between the generated image and its paired text. Consequently, prior pruning methods are less suitable for refining these synthetic datasets. Furthermore, conventional pruning approaches often rely on strict one-to-one mapping, discarding pairs that fall below a threshold. This can lead to the removal of potentially useful captions simply because their initial synthetic pairing was imperfect, even if a better match might exist within the dataset. To address these limitations, we introduce a flexible one-to-many mapping that reassigns well-formed captions to the most semantically aligned images. This enables more reliable synthetic image-text refinement and enhances the quality of training data for zero-shot captioning.

%%% 우리 방식은 synthetic data 에 적합한 - well construct 된 caption 을 가지고 잘못 페어링된 이미지를 re-assign 하는 방식. 여기서 strict one-to-one mapping 이 아니라 보다 flexible 하게 pair 를 구하는 one-to-many mapping 방식을 제안. 여기에 이어서 text 쪽 unimodality 만을 활용해 안정적으로 semantic alignment 를 계산하는 cycle-consistency based alignment score 를 제안함.

\section{Method}

\begin{figure*}[ht]
    \centering
       \includegraphics[width=1\textwidth]{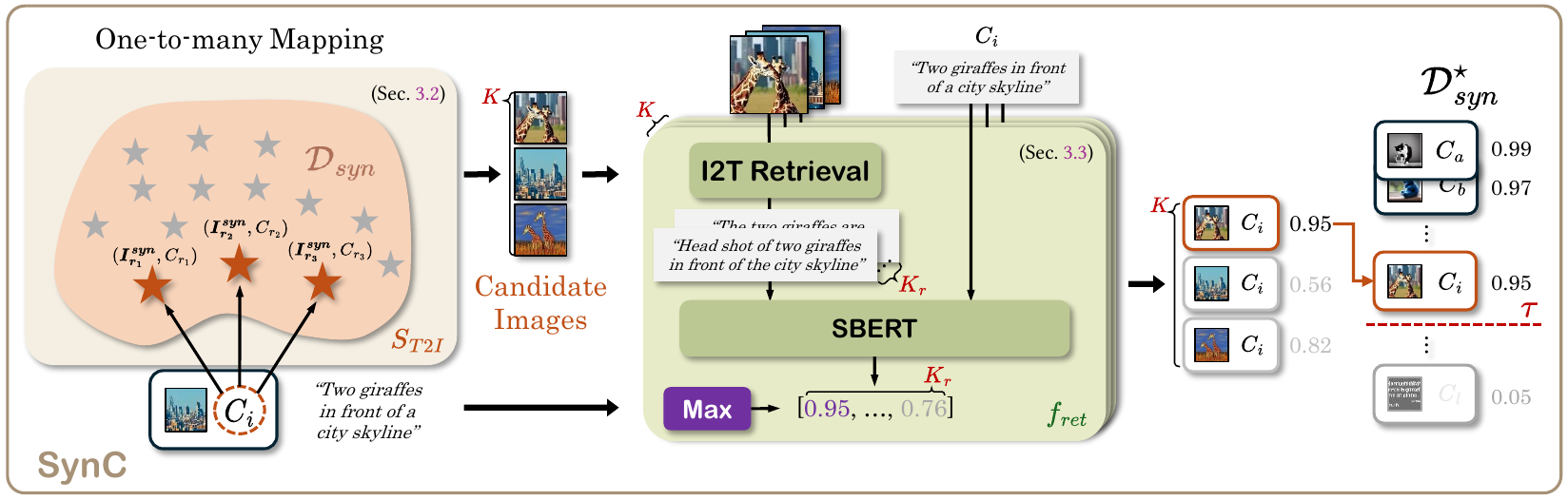}
    \caption{The overall pipeline of $\model$. $\model$ processes synthetic image caption pair dataset $\dsyn$ into refined $\dsyn^\star$. With selection function $\sonetomany\onevar$, $\model$ performs text-to-image retrieval to find relevant synthetic images with query caption for each caption $C_i \in \mathcal{C}$. With $K$ candidate images, $\model$ calculates the alignment score $\fret\twovar$ between $C_i$ and candidate images, picking the best pair and the similarity score. The $\model$ reassigns an adequate synthetic image if the generated image from $C_i$ does not fit with the input prompt. Iterating through captions $C_i$ in the textual corpus $\mathcal{C}$, we get a set of triplets consisting of a synthetic image, caption, and alignment score. Preserving upper $\tau$ fraction, $\model$ finally outputs a refined dataset $\dsyn^\star$.
    }
    \label{fig:pipeline}
\end{figure*}

% \subsection{Preliminary \djkim{(remove it?)}}\label{method3.0}

In this section, we detail our proposed method, $\model$, designed specifically to refine noisy synthetic datasets into well-aligned image-caption pairs. We begin by describing the basic concept of dataset pruning and the initial generation of a synthetic dataset $\dsyn$ (\Sref{method3.1}). Then, we introduce the core components of $\model$: a novel \textit{one-to-many} selection strategy $\sonetomany\onevar$ based on text-to-image retrieval (\Sref{method3.2}), and a tailored alignment scorer function $\fret\twovar$ 
%optimized 
for curating datasets intended for zero-shot image captioning (\Sref{method3.3}). The overall pipeline of $\model$ is depicted in \Fref{fig:pipeline}.

\subsection{Preliminary}\label{method3.1}

We first establish a general framework for dataset pruning. Our objective is to refine an initial image-caption dataset $\mathcal{D}=\{(I_i, C_i)\}_{i=1}^N$, comprising $N$ image-caption pairs, into a curated dataset $\mathcal{D}^\star$ characterized by higher semantic alignment and informativeness. The ultimate goal is to produce a dataset that enhances the training effectiveness of zero-shot image captioning models. This pruning process involves two key components: a \textit{selection function} $\s\onevar$ and a \textit{multi-modal scorer function} $\f\twovar$. 

The selection function $\s\onevar$ identifies a set of relevant candidate images for a given query caption. Formally, for a query caption $C$, $\s\onevar$ outputs a subset of $K$ candidate images from the image pool of dataset: 
%$S:C \rightarrow \{I_i\}_{i\in \text{cand}}$, 
$S(C)=\{I_i\}_{i\in \text{cand}}$, 
where $\text{cand}=\{i_1, i_2, ..., i_K\}$ is the set of indices for the selected images. This function maps a caption to one or more candidate images.

The multi-modal scorer function $\f\twovar$ quantifies the alignment between an image $I$ and a caption $C$. It takes an image-caption pair as input and produces a scalar similarity score $s=f(I, C)$. A higher score $s$ signifies stronger semantic correspondence between the image $I$ and the caption $C$ across different modalities. This scoring mechanism is crucial for identifying and retaining well-aligned image-caption pairs.

\noindent\textbf{Synthetic Dataset Generation.} 
Given an unlabeled text corpus $\mathcal{C}=\{C_i\}_{i=1}^N$ comprises of $N$ captions, we employ a text-to-image (T2I) generative model (e.g., SD~\cite{ldm}) to synthesize a corresponding set of images $\mathcal{I}_{syn}=\{I_i^{syn}\}_{i=1}^N$. we simply use the captions $C_i \in \mathcal{C}$ directly as input prompts without modification. Each synthetic image $I_i^{syn}$ is generated by feeding the corresponding caption $C_i$ to the T2I model. This process yields an initial, potentially noisy, paired synthetic dataset denoted as $\dsyn=\{(I_i^{syn},C_i)\}_{i=1}^N$.

\subsection{Synthetic Image Selection}\label{method3.2}
The initial synthetic dataset $\dsyn$ establishes a one-to-one correspondence between each caption $C_i$ and its generated image $I_i^{syn}$. This synthetic data assignment can be modeled by a simple selection function, $\sonetoone\onevar$, where a caption $C_i$ selects only its directly generated image:
\begin{equation}
    \sonetoone(C_i)=\{I_i^{syn}\}.
\end{equation}
This one-to-one approach is common in existing dataset pruning methods~\cite{clips, sieve, veclip, laclip} and T2I-based zero-shot image captioning models~\cite{pcm-net, icsd, syntic}.

However, T2I models can produce synthetic images that are irrelevant or fail to capture the full semantics of the input caption (\Fref{fig:teaser_one_vs_t2t}). Furthermore, simple captions may primarily describe dominant concepts while omitting finer details~\cite{icsd}. Conversely, human-annotated datasets like COCO~\cite{coco} and Flickr30k~\cite{flickr30k} often feature multiple reference captions per image.

Inspired by the one-to-many nature of human annotations and aiming to mitigate issues from imperfect T2I generation, we propose a novel \textit{one-to-many} selection strategy for synthetic image caption paired datasets, $\sonetomany\onevar$. Instead of being limited to the initially generated $I_i^{syn}$, our approach allows each caption $C_i$ to select from the pre-generated image pool $\mathcal{I}_{syn}$. Critically, while $\sonetoone\onevar$ discards misaligned pairs, $\sonetomany\onevar$ aims to actively find a better-aligned synthetic image for caption $C_i$ from the existing pool, thereby potentially refining valuable captions that would otherwise be discarded due to poor initial T2I generation.

Specifically, for each caption $C_i \in \mathcal{C}$, we perform text-to-image (T2I) retrieval over the entire set of generated synthetic images $\mathcal{I}_{syn}$. We use image encoder $\mathcal{E}_I(\cdot)$ and the text encoder $\mathcal{E}_T(\cdot)$ of a pre-trained vision-language model (e.g., CLIP~\cite{clip}, SigLIP~\cite{siglip, siglip2}) to find the $K$ synthetic images most relevant to the query caption $C_i$ in the pre-generated image pool. \textit{This selection process does not require additional image generation}; it simply retrieves from the existing image pool $\mathcal{I}_{syn}$. We define this one-to-many selection function, 
\begin{gather}
    \sonetomany(C_i)=\{I_r^{syn}\}_{r\in \rttot_i}, \\
    \text{where}~~\rttot_i=\operatorname*{Top-K}_{j \in [1,N]} \,
    \{\mathcal{E}_I(I_j^{syn}), \langle\mathcal{E}_T(C_i)\rangle\}.\label{eq:t2i_selection}
\end{gather}
%Here, $\langle \boldsymbol{x}, \boldsymbol{y}\rangle$ denotes the consine similarity $\cos(\boldsymbol{x}, \boldsymbol{y})$ 
Here, $\langle \cdot,\cdot \rangle$ denotes the consine similarity $\cos(\cdot,\cdot)$
and $\rttot_i=[r_1, ..., r_K]$ is the set of indices corresponding to the top-$K$ retrieved images for caption $C_i$. Thus, for each caption $C_i$, we consider $K$ candidate images $\{I_r^{syn}\}_{r\in \rttot_i}$. This forms a set of candidate pairs $\{(I_r^{syn}, C_i)\}_{r\in \rttot_i}$, from which the best-aligned pair will be chosen using our scoring function. Our proposed method $\model$ adopts $\s\onevar := \sonetomany\onevar$ as its selection function.

\subsection{Alignment Score Function}\label{method3.3}
A standard approach for scoring image-caption alignment is CLIPScore~\cite{clips}. Using CLIP's image encoder $\mathcal{E}^{CLIP}_I(\cdot)$ and text encoder $\mathcal{E}^{CLIP}_T(\cdot)$, the CLIPScore function $\fclip\twovar$ computes the cosine similarity between the embeddings of an image $I$ and a caption $C$:
\begin{gather}
    \fclip(I, C)=w\cdot\langle\mathcal{E}^{CLIP}_I(I), \mathcal{E}^{CLIP}_T(C)\rangle,\label{eq:t2i_retrieval}
\end{gather}
where $w$ is a scaling factor, we use $w=1.0$ for simplicity.

Our preliminary experiments indicate that using $\fclip\twovar$ for filtering yields marginal improvements for zero-shot image captioning models by removing some misaligned pairs. Notably, CLIP often prioritizes global alignment and may overlook fine-grained details or struggle with precise compositional understanding~\cite{sieve, aro, verbsinaction, oh2024preserving}.

To overcome these limitations, particularly in the context of refining noisy synthetic datasets, we introduce a novel \textit{retrieval-based multi-modal scorer function}, $\fret\twovar$. After T2I retrieval in the selection function, $\fret\twovar$ performs image-to-text (I2T) retrieval using the candidate synthetic image and then compares the retrieved captions with the original target caption within the unimodal text space. The underlying rationale is that the original caption $C_i$ often serves as a more reliable representation of the intended semantics than a potentially flawed synthetic image $I^{syn}$. Thus, rather than directly comparing image and text embeddings in a multi-modal space (as in CLIPScore), we leverage the consistency across retrieval directions. Inspired by cycle-consistency concepts~\cite{zhao2025doracycle, li2023leveraging}, our approach combines T2I retrieval for selection (in \Eref{eq:t2i_selection}) with I2T retrieval for scoring.

Specifically, for a candidate pair $(I^{syn}, C)$, we first use the synthetic image $I^{syn}$ to retrieve the most semantically similar captions from the original corpus $\mathcal{C}$ via I2T retrieval, using the same VLM encoders $\mathcal{E}_I$ and $\mathcal{E}_T$. We then measure the semantic similarity between the query caption $C$ and these image-retrieved captions. For this text-only comparison, we employ a dedicated unimodal text encoder, Sentence Transformer (SBERT)~\cite{sbert}, denoted as $\mathcal{E}_S(\cdot)$, known for its effectiveness in capturing sentence-level semantics. 

We define the retrieval-based scorer function $\fret\twovar$ as follows:
\begin{gather}
    \fret(I^{syn}, C)=\max_{r \in \ritot (I^{syn})}\langle\mathcal{E}_S(C_r), \mathcal{E}_S(C)\rangle, \\
    \text{where}~~\ritot(I^{syn})=\operatorname*{Top-K_r}_{j \in [1,N]} \, \{\langle\mathcal{E}_I(I^{syn}), \mathcal{E}_T(C_j)\rangle\}.
\end{gather}
Here, $\ritot(I^{syn})$ represents the set of top-$K_r$ captions retrieved from the corpus $\mathcal{C}$ based on their similarity to the synthetic image $I^{syn}$ in the VLM embedding space (using encoders $\mathcal{E}_I$ and $\mathcal{E}_T$ with \Eref{eq:t2i_retrieval}). The final score is the highest SBERT similarity between the target caption $C$ and the image-relevant captions. Our proposed method $\model$ uses $\f\twovar:=\fret\twovar$ as its scorer function.

With the synthetic image selection strategy $\s\onevar$ and scorer function $\f\twovar$, we iteratively calculate the alignment scores between $C_i$ and $I^{syn} \in S(C_i)$. The similarity $s_i$ is the highest alignment score,
\begin{gather}
    s_i=\max_{I^{syn} \in S(C_i)}f(I^{syn}, C_i), \\
    I_{i^\ast}^{syn}=\operatorname*{argmax}_{I^{syn} \in S(C_i)}f(I^{syn}, C_i),
\end{gather}
and we match the most related synthetic image $I_{i^{\ast}}^{syn}$ to $C_i$ and form a refined synthetic image caption pair $(I_{i^{\ast}}^{syn}, C_i)$.

The $\model$ gets an input $\dsyn$, iterating this procedure for each pair in $\dsyn$, outputs $\dsyn^\prime=\{(I_{i^{\ast}}^{syn}, C_i, s_i)\}_{i=1}^N$ which is a set of triplet consisting of refined synthetic image, caption, and alignment score $s_i$. Since some of the refined pairs may be misaligned due to the complexity of sentences or wrongly written captions~(e.g., grammar error, typo), $\model$ filters out the lower part of the $\dsyn^\prime$. After sorting $\dsyn^\prime$ by alignment score in descending order, $\model$ preserves the upper fraction, $\tau \in [0, 1]$. We get a refined synthetic image caption pair dataset $\dsyn^\star=\{(I_{i^{\ast}}^{syn}, C_i)\}_{i=1}^{\lfloor N\cdot \tau \rfloor}$ as a final output. We perform a floor operation $\lfloor \cdot \rfloor$ for integer casting $ N\cdot \tau$. The whole procedure of $\model$ can be written as
\begin{equation}
    \model(\dsyn, \tau) = \dsyn^\star.
\end{equation}

\section{Experiments}

\begin{table*}[t]
\caption{Overall results of in-domain captioning in MS-COCO and Flickr30k~\cite{coco, flickr30k}. We measure most commonly used captioning metrics BLEU@4~(B@4), METEOR~(M), ROUGE~(R), CIDEr~(C), and SPICE~(S)~\cite{cider, spice, bleu, meteor, rouge}. $\Baseline$ trained with synthetic dataset refined with $\model$ demonstrates a significant performance boost relative to the $\Baseline$ with a huge margin, achieving state-of-the-art performance. The best score is in \mff{bold}. $\dagger$ denotes the use of real images during training time.
}
\tabcolsep=5pt
\centering
\resizebox{0.8\linewidth}{!}{
\begin{tabular}{l|ccccc|ccccc}
\toprule[1.2pt]
\multirow{2}{*}{\textbf{Method}}  &  \multicolumn{5}{c|}{\textbf{COCO}} & \multicolumn{5}{c}{\textbf{Flickr30k}} \\
% & \multicolumn{1}{c}{B@4} & \multicolumn{1}{c}{M} & \multicolumn{1}{c}{C} & \multicolumn{1}{c|}{S} & \multicolumn{1}{c}{B@4} & \multicolumn{1}{c}{M} & \multicolumn{1}{c}{C} & \multicolumn{1}{c}{S} & \multicolumn{1}{c}{B@4} & \multicolumn{1}{c}{M} & \multicolumn{1}{c}{C} & \multicolumn{1}{c}{S} \\
& B@4 & M & R & C & S & B@4 & M & R & C & S \\
\midrule
\multicolumn{11}{l}{\textcolor{\VariantColor}{\textit{ViT-B/32}}} \\
% CapDec \cite{capdec} \ven{EMNLP'22}               & RN50x4 & 26.4 & 25.1 & 51.8 & 91.8 & -- & 17.7 & 20.0 & 43.9 & 39.1 & -- \\
DeCap \cite{decap} \ven{ICLR'23}                  & 24.7 & 25.0 & -- & 91.2 & 18.7 & 21.2 & 21.8 & -- & 56.7 & 15.2 \\
ViECap \cite{viecap} \ven{ICCV'23}                & 27.2 & 24.8 & -- & 92.9 & 18.2 & 21.4 & 20.1 & -- & 47.9 & 13.6 \\
ICSD \cite{icsd} \ven{AAAI'24}                    & 29.9 & 25.4 & -- & 96.6 & -- & 25.2 & 20.6 & -- & 54.3 & -- \\
SynTIC \cite{syntic} \ven{AAAI'24}                & 29.9 & 25.8 & 53.2 & 101.1 & 19.3 & 22.3 & 22.4 & 47.3 & 56.6 & 16.6 \\
IFCap \cite{ifcap} \ven{EMNLP'24}                 & 30.8 & \mff{26.7} & -- & 108.0 & 20.3 & 23.5 & 23.0 & -- & 64.4 & 17.0 \\
$\Baseline$ \cite{pcm-net} \ven{ECCV'24}          & 31.5 & 25.9 & 53.9 & 103.8 & 19.7 & 26.9 & 23.0 & 50.1 & 61.3 & 16.8 \\
\rowcolor{\lineColor}
$+\model$                                & \mff{33.6} & \mff{26.7} & \mff{55.3} & \mff{112.0} & \mff{20.5} & \mff{28.2} & \mff{23.4} & \mff{50.5} & \mff{65.8} & \mff{17.1} \\
% CapDec \cite{capdec} \ven{EMNLP'22}             & RN50x4   & 26.4 & 25.1 & 51.8 & 91.8  & --   & 17.7 & 20.0 & 43.9 & 39.1 & -- \\
% DeCap \cite{decap} \ven{ICLR'23}                & ViT-B/32 & 24.7 & 25.0 & -- & 91.2  & 18.7 & 21.2 & 21.8 & -- & 56.7 & 15.2 \\
% ViECap \cite{viecap} \ven{ICCV'23}              & ViT-B/32 & 27.2 & 24.8 & -- & 92.9  & 18.2 & 21.4 & 20.1 & -- & 47.9 & 13.6 \\
% ICSD$^\dagger$ \cite{icsd} \ven{AAAI'24}        & ViT-B/32 & 29.9 & 25.4 & -- & 96.6  & --   & 25.2 & 20.6 & -- & 54.3 & --   \\
% SynTIC$^\dagger$\cite{syntic} \ven{AAAI'24}     & ViT-B/32 & 29.9 & 25.8 & 53.2 & 101.1 & 19.3 & 22.3 & 22.4 & 47.3 & 56.6 & 16.6 \\
% IFCap \cite{ifcap} \ven{EMNLP'24}               & ViT-B/32 & 30.8 & 26.7 & -- & 108.0 & 20.3 & 23.5 & 23.0 & -- & 64.4 & 17.0 \\
% \midrule
% \rowcolor{\lineColor}
% $\Baseline$$^\dagger$\cite{pcm-net} \ven{ECCV'24}  & ViT-B/32 & 31.5 & 25.9 & 53.9 & 103.8 & 19.7 & 26.9 & 23.0 & 50.1 & 61.3 & 16.8 \\
% $+\model$ (T2I) &  & 33.6 & 26.7 & 55.3 & 112.0 & 20.5 & 27.8 & 23.1 & 50.5 & 65.3 & 16.8 \\
% $+\model$ (T2T) &  & 32.8 & 26.2 & 54.6 & 108.4 & 20.3 & 26.9 & 23.0 & 50.2 & 63.1 & 16.4 \\

\midrule 
\multicolumn{11}{l}{\textcolor{\VariantColor}{\textit{ViT-L/14}}} \\
CgT-GAN$^\dagger$ \cite{cgtgan} \ven{ACMMM'23}  & 30.3 & 26.9 & 54.5 & 108.1 & 20.5 & 24.1 & 22.6 & 48.2 & 64.9 & 16.1 \\
$\Baseline$ \cite{pcm-net} \ven{ECCV'24}        & 33.6 & 26.9 & 55.4 & 113.6 & 20.8 & 28.5 & 24.3 & 51.4 & 69.5 & 18.2 \\
\rowcolor{\lineColor}
$+\model$                              & \mff{35.2} & \mff{27.8} & \mff{56.5} & \mff{119.8} & \mff{21.9} & \mff{29.6} & \mff{24.8} & \mff{52.5} & \mff{75.6} & \mff{18.5} \\
% CgT-GAN \ven{ACMMM'23} & ViT-L/14 & 30.3 & 26.9 & 54.5 & 108.1 & 20.5 & 24.1 & 22.6 & 48.2 & 64.9 & 16.1 \\
% \midrule
% \rowcolor{\lineColor}
% $\Baseline$$^\dagger$\cite{pcm-net} \ven{ECCV'24}  & ViT-L/14 & 33.6 & 26.9 & 55.4 & 113.6 & 20.8 & 28.5 & 24.3 & 51.4 & 69.5 & 18.2 \\
% $+\model$ (T2I) &  & 35.2 & 27.8 & 56.5 & 119.8 & 21.9 & 29.6 & 24.8 & 52.5 & 75.6 & 18.5  \\
\bottomrule[1.2pt]
\end{tabular}
}\label{tb:mm_indomain}
\end{table*}
\begin{table}[t]
\caption{Results on the cross-domain captioning. $\model$ improves $\Baseline$ with achieving state-of-the-art performance in most metrics.}
\tabcolsep=2pt
\resizebox{\columnwidth}{!}{
\begin{tabular}{l|ccccc|ccccc}
\toprule[1.2pt]
\multirow{2}{*}{\textbf{Method}} & \multicolumn{5}{c|}{\textbf{COCO $\Longrightarrow$ Flickr30k}} & \multicolumn{5}{c}{\textbf{Flickr30k $\Longrightarrow$ COCO}} \\
& B@4 & M & R & C & S & B@4 & M & R & C & S \\
\midrule
\multicolumn{11}{l}{\textcolor{\VariantColor}{\textit{ViT-B/32}}} \\
% CapDec \cite{capdec} & RN50x4 & 17.3 & 18.6 & 42.7 & 35.7 & -- & 9.2 & 16.3 & 36.7 & 27.3 & -- \\
DeCap \cite{decap}           & 16.3 & 17.9 & -- & 35.7 & 11.1 & 12.1 & 18.0 & -- & 44.4 & 10.9 \\
ViECap \cite{viecap}         & 17.4 & 18.0 & -- & 38.4 & 11.2 & 12.6 & 19.3 & -- & 54.2 & 12.5 \\
SynTIC \cite{syntic}         & 17.9 & 18.6 & 42.7 & 38.4 & 11.9 & 14.6 & 19.4 & 40.9 & 47.0 & 11.9 \\
IFCap \cite{ifcap}           & 17.8 & 19.4 & -- & 47.5 & 12.7 & 14.7 & \mff{20.4} & -- & \mff{60.7} & \mff{13.6} \\
$\Baseline$ \cite{pcm-net}   & 20.8 & 19.2 & 45.2 & 45.5 & 12.9 & 17.1 & 19.6 & 43.2 & 54.9 & 12.8 \\
\rowcolor{\lineColor}
+$\model$         & \mff{22.4} & \mff{20.0} & \mff{45.9} & \mff{49.5} & \mff{13.4} & \mff{18.4} & 20.0 & \mff{44.1} & 58.4 & 13.4 \\
% CapDec \cite{capdec}  & RN50x4 & 17.3 & 18.6 & 42.7 & 35.7 & -- & 9.2 & 16.3 & 36.7 & 27.3 & -- \\
% DeCap \cite{decap}    & ViT-B/32 & 16.3 & 17.9 & -- & 35.7 & 11.1 & 12.1 & 18.0 & -- & 44.4 & 10.9 \\
% ViECap \cite{viecap}  & ViT-B/32 & 17.4 & 18.0 & -- & 38.4 & 11.2 & 12.6 & 19.3 & -- & 54.2 & 12.5 \\
% SynTIC \cite{syntic}  & ViT-B/32 & 17.9 & 18.6 & 42.7 & 38.4 & 11.9 & 14.6 & 19.4 & 40.9 & 47.0 & 11.9 \\
% IFCap \cite{ifcap}    & ViT-B/32 & 17.8 & 19.4 & -- & 47.5 & 12.7 & 14.7 & 20.4 & -- & 60.7 & 13.6 \\
% \rowcolor{\lineColor}
% $\Baseline$\cite{pcm-net} & ViT-B/32 & 20.8 & 19.2 & 45.2 & 45.5 & 12.9 & 17.1 & 19.6 & 43.2 & 54.9 & 12.8 \\
% +$\model$ (T2I)       & ViT-B/32 & 22.4 & 20.0 & 45.9 & 49.5 & 13.4 & 18.5 & 19.8 & 44.0 & 57.6 & 13.4 \\
%+$\model$             & ViT-B/32 & 22.7 & 19.8 & 45.7 & 49.6 & 13.5 & 18.0 & 19.6 & 43.6 & 56.2 & 12.8 \\

\midrule
\multicolumn{11}{l}{\textcolor{\VariantColor}{\textit{ViT-L/14}}} \\
CgT-GAN$^\dagger$ \cite{cgtgan}      & 17.3 & 19.6 & 43.9 & 47.5 & 12.9 & 15.2 & 19.4 & 40.9 & 58.7 & 13.4\\

$\Baseline$ \cite{pcm-net} & 23.9 & 21.2 & 47.8 & 55.9 & 14.2 & 17.9 & 20.3 & 44.0 & 61.3 & 13.5 \\
\rowcolor{\lineColor}
+$\model$       & \mff{25.2} & \mff{22.1} & \mff{49.6} & \mff{60.3} & \mff{15.8} & \mff{19.9} & \mff{21.3} & \mff{45.6} & \mff{66.8} & \mff{14.9} \\
% CgT-GAN \cite{cgtgan}       & 17.3 & 19.6 & 43.9 & 47.5 & 12.9 & 15.2 & 19.4 & 40.9 & 58.7 & 13.4\\
% \rowcolor{\lineColor}
% $\Baseline$           & ViT-L/14 & 23.9 & 21.2 & 47.8 & 55.9 & 14.2 & 17.9 & 20.3 & 44.0 & 61.3 & 13.5 \\
% +$\model$ (T2I)       & ViT-L/14 & 25.2 & 22.1 & 49.6 & 60.3 & 15.8 & 19.9 & 21.3 & 45.6 & 66.8 & 14.9 \\
%+$\model$             & ViT-L/14 & 24.3 & 21.5 & 49.1 & 60.5 & 15.2 & 19.5 & 20.9 & 45.3 & 65.3 & 14.1 \\ 
\bottomrule[1.2pt]
\end{tabular}
}\label{tb:mm_crossdomain}
\end{table}

\subsection{Implementation Details and Settings}

We adopt PCM-Net~\cite{pcm-net} as the baseline zero-shot image captioning model. While several related methods exist~\cite{icsd, syntic, pcm-net}, PCM-Net is the only publicly available implementation, enabling reproducible comparison. We evaluate our synthetic data refinement method, $\model$, by applying it to the dataset used by the open-sourced PCM-Net~\cite{pcm-net} zero-shot captioning baseline. For MS-COCO~\cite{coco} and Flickr30k~\cite{flickr30k} evaluations, we refine PCM-Net's public synthetic dataset $\mathcal{D}_\text{SynthImgCap}$~\cite{pcm-net}.

For experiments on CC3M~\cite{CC3M} and SS1M~\cite{SS1M}, we utilize stable diffusion v1.4\footnote{\href{https://huggingface.co/CompVis/stable-diffusion-v1-4}{https://huggingface.co/CompVis/stable-diffusion-v1-4}} from huggingface~\cite{wolf2020transformers} to generate synthetic images with caption source of CC3M and SS1M. We set 512$\times$512 as an image resolution, 20 sampling steps with DPMSolver~\cite{lu2022dpm}.
 
Our method employs $\s\onevar := \sonetomany\onevar$ selection function using SigLIP2 ViT-B/16@256~\cite{siglip2} for retrieval with $K=15$ candidates images based on \Fref{fig:mm_ablation}. The scorer function $\f\twovar := \fret\twovar$ combines SigLIP2 ViT-B/16@256 and a lightweight distilled sentence transformer, all-MiniLM-L6v2~\cite{wang2020minilm}. We select preserving ratio $\tau=0.9$ , as in \Fref{fig:mm_reserve_ratio_pcmnet} and $K_r=2$ based on \Fref{fig:mm_ablation}.

\begin{table}[t]
\caption{Result on NoCaps~\cite{nocaps} validation split under the cross-domain setting. $\model$ demonstrates performance enhancement in out-of-domain scenario.}
\tabcolsep=4pt
\resizebox{0.8\columnwidth}{!}{
\begin{tabular}{l|cc|cc|cc|cc}
\toprule[1.2pt]
\multirow{3}{*}{\textbf{Method}}  & \multicolumn{8}{c}{\textbf{NoCaps}} \\
 & \multicolumn{2}{c}{In} & \multicolumn{2}{c}{Near} & \multicolumn{2}{c}{Out} & \multicolumn{2}{c}{Entire} \\
 & C & S & C & S & C & S & C & S \\
% & B@4 & M & C & S & B@4 & M & C & S \\
\midrule
ViECap \cite{viecap} & 61.1 & 10.4 & 64.3 & 9.9 & 65.0 & 8.6 & 66.2 & 9.5 \\
SYN-ViECap \cite{pcm-net} & 61.7 & \mff{10.5} & 68.5 & 10.3 & 71.4 & \mff{9.4} & 70.5 & 10.0 \\
% SYN-ViECap$^\text{Reproduced}$ & 54.7 & 9.6 & 63.3 & 9.4 & 68.2 & 8.6 & 65.9 & 9.2 \\
\rowcolor{\lineColor}
+\model & \mff{65.7} & 10.4 & \mff{71.2} & \mff{10.5} & \mff{71.6} & \mff{9.4} & \mff{72.7} & \mff{10.1} \\
% ViECap \cite{viecap}      &  61.1 & 10.4 & 64.3 & 9.9  & 65.0 & 8.6 & 66.2 & 9.5 \\
% SYN-ViECap \cite{pcm-net} &  61.7 & 10.5 & 68.5 & 10.3 & 71.4 & 9.4 & 70.5 & 10.0 \\
% SYN-ViECap$^\text{Reproduced}$ &  54.7 & 9.6 & 63.3 & 9.4 & 68.2 & 8.6 & 65.9 & 9.2 \\
% SYN-ViECap + \model       &  65.7 & 10.4 & 71.2 & 10.5 & 71.6 & 9.4 & 72.7 & 10.1 \\
\bottomrule[1.2pt]
\end{tabular}
}\label{tb:mm_nocaps}
\end{table}
  
\begin{table}
\caption{Result on cross-domain image captioning with 200k synthetic image caption pairs generated from text source of CC3M~\cite{CC3M} and SS1M~\cite{SS1M} respectively. In every setting, $\model$ demonstrates consistent performance enhancement.}
\tabcolsep=4pt
\centering
\resizebox{0.8\linewidth}{!}{
\begin{tabular}{l|ccccc}
\toprule[1.2pt]
\multirow{2}{*}{\textbf{Dataset}} & \multicolumn{5}{c}{\textbf{Dataset $\Longrightarrow$ COCO}}  \\
& B@4 & M & R & C & S \\
\midrule
% \multicolumn{6}{l}{\textbf{Other zero-shot image captioning models}} \\
% DeCap \cite{decap}            & CC3M & 8.8  & 16.0 & -- & 42.1 & 10.9  \\
% DeCap                         & SS1M & 8.9  & 17.5 & -- & 50.6 & 13.1  \\
% ICSD$^\dagger$ \cite{icsd}    & SS1M & 13.6 & 18.3 & -- & 54.2 & --   \\
% SynTIC$^\dagger$\cite{syntic} & SS1M & 13.3 & 17.6 & -- & 55.6 & 13.3  \\
% \multicolumn{1}{l}{\Baseline \textcolor{\VariantColor} {\textit{ViT-L/14}}~\cite{pcm-net}} \\
\multicolumn{1}{l}{\Baseline{} \textcolor{\VariantColor}{\textit{ViT-L/14}}~\cite{pcm-net}} \\
$\mathcal{D}_\text{CC3M}$ & 9.6 & 14.0 & 34.7 & 41.3 & 10.1 \\
\rowcolor{\lineColor}
$\plusmodel$ & \mff{11.0} & \mff{15.3} & \mff{36.9} & \mff{47.7} & \mff{11.3} \\
% $\mathcal{D}_\text{CC3M}$       & 9.6 & 14.0 & 34.7 & 41.3 & 10.1 \\
% $\mathcal{D}_\text{CC3M}^\star$ & 11.0 & 15.3 & 36.9 & 47.7 & 11.3 \\
\midrule
$\mathcal{D}_\text{SS1M}$ & 10.7 & 13.4 & 29.9 & 47.5 & \mff{11.3} \\
\rowcolor{\lineColor}
$\plusmodel$ & \mff{11.3} & \mff{14.2} & \mff{30.9} & \mff{49.1} & \mff{11.3} \\
% $\mathcal{D}_\text{SS1M}$       & 10.7 & 13.4 & 29.9 & 47.5 & 11.3 \\
% $\mathcal{D}_\text{SS1M}^\star$ & 11.3 & 14.2 & 30.9 & 49.1 & 11.3 \\
\bottomrule[1.2pt]
\end{tabular}
}\label{tb:mm_cc3m}
\end{table}

Evaluation uses MS-COCO and Flickr30k (Karpathy split~\cite{karpathy}) for in-domain tasks and the NoCaps~\cite{nocaps} validation set for out-of-domain generalization. We report standard metrics: BLEU@4~\cite{bleu}, METEOR~\cite{meteor}, ROUGE-L~\cite{rouge}, CIDEr~\cite{cider}, and SPICE~\cite{spice}.

\subsection{Main Results}

We evaluate the effectiveness of $\model$ by comparing the performance of the baseline PCM-Net model~\cite{pcm-net} when trained on its original synthetic dataset $\mathcal{D}_\text{SynthImgCap}$ versus the dataset refined by our method $\mathcal{D}_\text{SynthImgCap}^\star$. We denote the baseline as $\Baseline$ and the baseline trained with our refined dataset as $\plusmodel$.

\noindent\textbf{Comparison with State-of-the-Art Methods.} We compare $\plusmodel$ against contemporary state-of-the-art zero-shot image captioning models. These include methods trained solely on text data like DeCap~\cite{decap}, ViECap~\cite{viecap}, and IFCap~\cite{ifcap}, as well as methods utilizing synthetic images like the original PCM-Net~\cite{pcm-net}, SynTIC~\cite{syntic}, ICSD~\cite{icsd}, and CgT-GAN~\cite{cgtgan}. Following common practice, we group results based on the backbone vision transformer: CLIP~\cite{clip} ViT-B/32 and ViT-L/14.

\begin{figure*}[t]
    \centering    
    \includegraphics[width=0.9\textwidth]{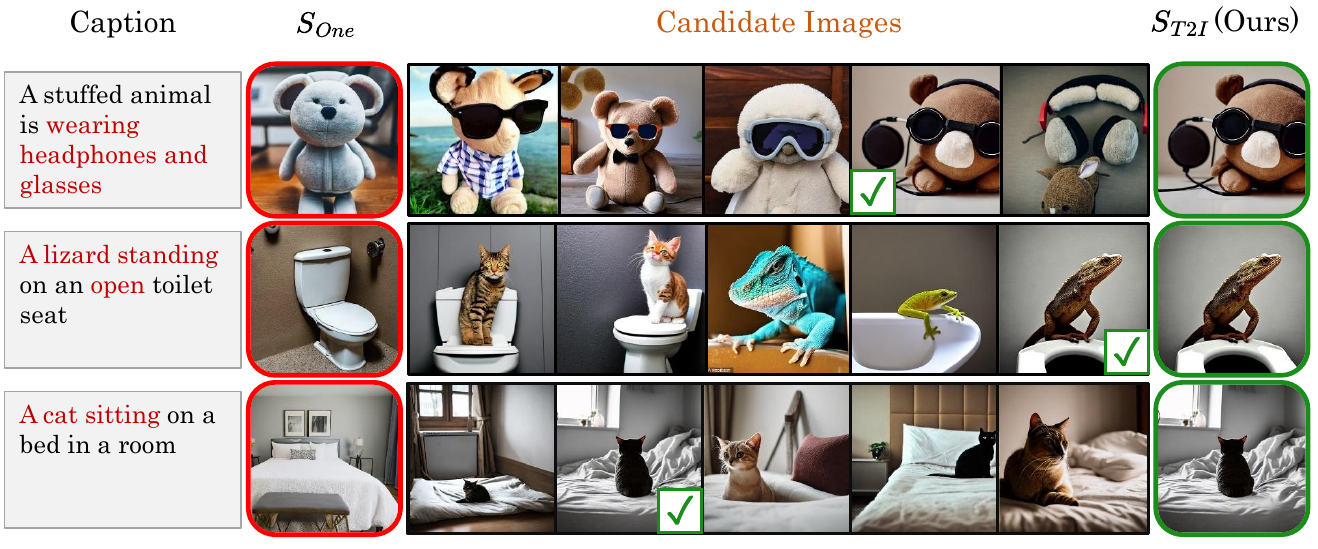}
    \caption{Qualitative results on comparison between $\sonetoone\onevar$ and our $\sonetomany\onevar$. $\sonetomany\onevar$ can search for an appropriate synthetic image for a given query caption. In combination with $\fret\twovar$, $\model$ can refine a misaligned pair while $\sonetoone\onevar$ discards the pair.}
    \label{fig:mm_qualitative}
\end{figure*}

\begin{table}[t]
\caption{Comparison $\model$ with web data pruning methods. Augmented-based pruning methods degrade $\Baseline$ performance. Metric-based turning methods marginally improve $\Baseline$, and adding our proposed $\sonetomany\onevar$ demonstrates additional performance boost.}
\tabcolsep=4pt
\resizebox{0.8\columnwidth}{!}{
\begin{tabular}{l|ccccc}
\toprule[1.2pt]
\multirow{2}{*}{\textbf{Method}} &  \multicolumn{5}{c}{\textbf{COCO}} \\
 & B@4 & M & R & C & S \\
% & B@4 & M & C & S & B@4 & M & C & S \\
\midrule
$\Baseline$ \cite{pcm-net} & 31.5 & 25.9 & 53.9 & 103.8 & 19.7 \\
\rowcolor{\lineColor}
+$\model$ & \mff{33.6} & \mff{26.7} & \mff{55.3} & \mff{112.0} & \mff{20.5} \\
\midrule
\multicolumn{6}{l}{\textbf{Augment-based pruning}} \\
VeCLIP        \cite{veclip} \ven{ECCV'24} & 27.0 & 25.9 & 52.5 & 93.9 & 19.8 \\
LaCLIP        \cite{laclip} \ven{NeurIPS'23} & 23.7 & 22.6 & 49.6 & 80.5 & 16.3 \\
Recaptioning  \cite{synthclip} \ven{ArXiv'24} & 27.8 & 25.8 & 52.8 & 95.1 & 19.2 \\
\midrule
\multicolumn{6}{l}{\textbf{Metric-based pruning}} \\
Sieve         \cite{sieve} \ven{CVPR'24} & 32.4 & 25.8 & 54.1 & 106.3 & \mff{19.9} \\
\rowcolor{\lineColor}
Sieve         + $\sonetomany$ & \mff{32.9} & \mff{26.1} & \mff{54.6} & \mff{108.1} & \mff{19.9} \\
% Sieve         \cite{sieve} \ven{CVPR'24} & 32.4 & 25.8 & 54.1 & 106.3 & 19.9 \\
% \rowcolor{\lineColor}
% Sieve         + $\sonetomany$     & 32.9 & 26.1 & 54.6 & 108.1 & 19.9 \\
CLIPScore     \cite{clips} \ven{EMNLP'21} & 31.6 & 25.8 & 53.7 & 104.4 & 19.5 \\
\rowcolor{\lineColor}
CLIPScore     + $\sonetomany$ & \mff{32.3} & \mff{26.0} & \mff{54.3} & \mff{106.5} & \mff{19.8} \\
% CLIPScore     \cite{clips} \ven{EMNLP'21} & 31.6 & 25.8 & 53.7 & 104.4 & 19.5 \\
% \rowcolor{\lineColor}
% CLIPScore     + $\sonetomany$    & 32.3 & 26.0 & 54.3 & 106.5 & 19.8 \\

% $\Baseline$ \cite{pcm-net} & 31.5 & 25.9 & 53.9 & 103.8 & 19.7 \\
% Sieve         \cite{sieve} \ven{CVPR'24}        & 32.4 & 25.8 & 54.1 & 106.3 & 19.9 \\
% Recaptioning  \cite{synthclip} \ven{ArXiv'24}   & 27.8 & 25.8 & 52.8 & 95.1 & 19.2 \\
% VeCLIP        \cite{veclip} \ven{ECCV'24}       & 27.0 & 25.9 & 52.5 &  93.9 & 19.8 \\
% LaCLIP        \cite{laclip} \ven{NeurIPS'23}    & 23.7 & 22.6 & 49.6 & 80.5 & 16.3 \\
% CLIPScore     \cite{clips} \ven{EMNLP'21}       & 32.3 & 25.7 & 54.1 & 104.7 & 19.5 \\
% $\model$                                        & 33.6 & 26.7 & 55.3 & 112.0 & 20.5 \\
\bottomrule[1.2pt]
\end{tabular}
}\label{tb:mm_webdatafiltering}
\end{table}

\noindent\textbf{In-domain captioning.} The results are presented in \Tref{tb:mm_indomain}. Applying our $\model$ yields substantial improvements over the $\Baseline$ across \textit{all} evaluation metrics on both datasets and for both backbones. Notably, on MS-COCO, $\plusmodel$ achieves CIDEr gains of \textbf{+8.2} (ViT-B/32) and \textbf{+6.2} (ViT-L/14) over the $\Baseline$. Significant CIDEr improvements are also observed on Flickr30k, with gains of \textbf{+4.5} (ViT-B/32) and \textbf{+6.2} (ViT-L/14). These consistent gains verify that refining the synthetic image-caption pairs with $\model$ markedly enhances the performance of the zero-shot captioning model.

\noindent\textbf{Cross-Domain and Out-of-Domain Generalization.} We further evaluate $\model$ in cross-domain and out-of-domain settings. As shown in \Tref{tb:mm_crossdomain}, $\plusmodel$ consistently outperforms the $\Baseline$ across all metrics in cross-dataset transfers. It achieves state-of-the-art results for COCO$\rightarrow$Flickr30k (both ViT-B/32 and ViT-L/14 backbones) and for Flickr30k$\rightarrow$COCO (ViT-L/14). For Flickr30k$\rightarrow$COCO (ViT-B/32), it also surpasses the baseline significantly and delivers strong competitive performance against prior art.

On the NoCaps validation set (\Tref{tb:mm_nocaps}), we use SYN-ViECap~\cite{viecap}, trained on $\mathcal{D}_\text{SynthImgCap}$. SYN-ViECap trained on the refined dataset by $\model$ demonstrates enhanced out-of-domain generalization, improving CIDEr across all splits and SPICE on most splits compared to the baseline performance reported in~\cite{pcm-net}. These results confirm the broad applicability and generalization benefits of refining synthetic data with $\model$.

\begin{table}[t]
\caption{Computational cost measure on web dataset pruning methods and $\model$. Cost is represented by RTX A6000 GPU hours. $\Delta$ represents the CIDEr score gain in the MS-COCO test split compared to the $\Baseline$. }
\tabcolsep=4pt
\centering
\resizebox{0.8\columnwidth}{!}{
\begin{tabular}{l|cccc|c|c}
\toprule[1.2pt]
\textbf{Method}               & Preprocess & $\mathcal{E(\cdot)}$ & $S$  & $f$  & Total & $\Delta$ \\
\midrule
VeCLIP    \cite{veclip}       &      46.5h & --        &  --     &  --  & 46.5h & \mdd{-9.9}  \\
LaCLIP    \cite{laclip}       &        28h & --        &  --     &  --  & 28h   & \mdd{-23.3} \\
Recaptioning \cite{synthclip} & 36h        & --        &  --     &  --  & 36h   & \mdd{-8.7} \\
% Sieve     \cite{sieve}        &        36h & 34m 53s + 8m 30s    &  --     &  4s  & 36h 43m 27s \\
Sieve     \cite{sieve}        &        36h & 0.7h      &  --     &  0h  & 36.7h & \muu{+2.5} \\

% CLIPScore \cite{clips}        & --         & 51m 30s   &  --     &   2s                &  51m 32s       \\
CLIPScore \cite{clips}        & --         & 0.9h   &  --     &   0h  &  0.9h   & \muu{+0.6} \\
\rowcolor{\lineColor}
%$\model$                      & -- & 1h 23m + 7m 20s + 2m 3s  & 22m 40s  & 22m 20s + 1m 46s     &  1h 13m 50s    \\
$\model$                      & -- & 1.5h  & 0.4h  & 0.4h     &  2.3h           & \muu{+8.2} \\
\bottomrule[1.2pt]
\end{tabular}
}\label{tb:mm_webdatafilteringtime}
\end{table}

\noindent\textbf{Cross-domain Captioning with Web Data.} We further verify our proposed $\model$ gain benefit from human annotated captions. Since COCO and Flickr30k consist of multiple captions per image, the annotation style could affect the performance enhancement of $\model$. To model the sparse environment, we use a part of CC3M and SS1M~\cite{SS1M, CC3M} training captions to create synthetic image caption datasets $\mathcal{D}_\text{CC3M}$ and $\mathcal{D}_\text{SS1M}$. We randomly sample 200,000 captions from each dataset, then perform image generation as illustrated in \Sref{method3.1}. We train $\Baseline$ with ViT-L/14 with $\mathcal{D}_\text{CC3M}$ and $\mathcal{D}_\text{SS1M}$ respectively, then test on COCO test split, the result of $\model$ on synthetic image caption paired datasets with the text source from web crawled datasets is in \Tref{tb:mm_cc3m}. $\model$ successfully enhances the performance of $\Baseline$ in most metrics. This verifies that $\model$ can refine in a sparse caption environment as well.

\subsection{Comparison with other Pruning Methods}

\noindent\textbf{Comparison with Web Data Filtering Methods.} We compare $\model$ against standard web data filtering methods applied to synthetic dataset $\mathcal{D}_\text{SynthImgCap}$. These include augment-based methods (VeCLIP~\cite{veclip}, LaCLIP~\cite{laclip}, and Recaptioning~\cite{synthclip}) and metric-based methods (Sieve~\cite{sieve} and CLIPScore~\cite{clips}). As shown in \Tref{tb:mm_webdatafiltering}, when applied to our baseline model:
1) \textit{Augment-based methods}, designed for textual noise, degrade performance likely because the primary noise source in synthetic data is visual, not textual. 2) Metric-based methods (Sieve, CLIPScore) offer only marginal improvements over the baseline, significantly less effective than our $\plusmodel$ approach, highlighting the limitations of their strict one-to-one pair filtering.

However, our core one-to-many selection strategy ($\sonetomany\onevar$, in \Tref{tb:mm_webdatafiltering}) complements existing metric-based filters. Applying $\sonetomany\onevar$ after initial filtering with Sieve or CLIPScore yields further consistent improvements across most metrics (e.g., CIDEr gains of +1.8 and +2.1, respectively). This demonstrates the added value and compatibility of our selection strategy.

\begin{table}[t]
\caption{$\model$ on different zero-shot image captioning models. $\model$ consistently exhibits performance enhancement.}
\tabcolsep=4pt
\resizebox{0.8\columnwidth}{!}{
\begin{tabular}{l|ccccc}
\toprule[1.2pt]
\multirow{2}{*}{\textbf{Method}} &  \multicolumn{5}{c}{\textbf{COCO}} \\
 & B@4 & M & R & C & S \\
\midrule
CapDec \cite{capdec} & 26.4 & 25.1 & 51.8 & 91.8 & -- \\
$\mathcal{D}_\text{SynthImgCap}$ & 25.4 & 24.7 & 50.4 & 89.3 & 18.6 \\
\rowcolor{\lineColor}
+\model & \mff{27.7} & \mff{25.7} & \mff{52.1} & \mff{96.4} & \mff{19.3} \\
% CapDec \cite{capdec}    & 26.4 & 25.1 & 51.8 & 91.8 & -- \\
% \noFiltering            & 25.4 & 24.7 & 50.4 & 89.3 & 18.6 \\
% \model                  & 27.7 & 25.7 & 52.1 & 96.4 & 19.3 \\
\midrule
ViECap   \cite{viecap} & 27.2 & \mff{24.8} & -- & 92.9 & 18.2 \\
$\mathcal{D}_\text{SynthImgCap}$ & 26.3 & 24.1 & 49.9 & 91.4 & 17.9 \\
\rowcolor{\lineColor}
+\model & \mff{27.7} & \mff{24.8} & \mff{51.6} & \mff{96.4} & \mff{18.5} \\
% ViECap   \cite{viecap}  & 27.2 & 24.8 & -- & 92.9 & 18.2 \\
% \noFiltering            & 26.3 & 24.1 & 49.9 & 91.4 & 17.9 \\
% \model                  & 27.7 & 24.8 & 51.6 & 96.4 & 18.5 \\
% \midrule
% \rowcolor{\lineColor}
% Knight \cite{knight}  & 27.8 & 26.4 & 52.3 & 98.9  & 19.6 \\
% \noFiltering          & 29.5 & 26.8 & 53.7 & 104.8 & 20.5 \\
% \model                & 29.1 & 26.6 & 53.2 & 103.5 & 20.4 \\
\midrule
IFCap    \cite{ifcap} & 30.8 & \mff{26.7} & -- & 108.0 & \mff{20.3} \\
$\mathcal{D}_\text{SynthImgCap}$ & 29.9 & 25.7 & 53.0 & 106.3 & 19.6 \\
\rowcolor{\lineColor}
+$\model$ & \mff{31.1} & 26.4 & \mff{53.6} & \mff{108.6} & \mff{20.3} \\
% IFCap    \cite{ifcap} & 30.8 & 26.7 & -- & 108.0 & 20.3 \\
% \noFiltering          & 29.9 & 25.7 & 53.0 & 106.3 & 19.6 \\
% $\model$              & 31.1 & 26.4 & 53.6 & 108.6 & 20.3 \\
\bottomrule[1.2pt]
\end{tabular}
}\label{tb:mm_zsic+ours}
\end{table}

\noindent\textbf{Computational Cost.} We measure computational cost in RTX A6000 GPU hours. Our analysis considers four stages potentially involved in pruning methods: \textbf{1)} preprocessing (e.g., LLM inference and image captioning), \textbf{2)} embedding generation $\mathcal{E}(\cdot)$ using VLM encoders (e.g., CLIP~\cite{clip} and SigLIP~\cite{siglip, siglip2}) or unimodal text encoders (e.g., Sentence Transformer~\cite{wang2020minilm}), \textbf{3)} the selection strategy $\s\onevar$, and \textbf{4)} calculation of alignment scores using a multimodal scorer function $\f\twovar$.

For $\model$, it requires two embedding steps: generating SigLIP2 embeddings for the full synthetic dataset ($\sim$1.5 hours) and Sentence Transformer embeddings for the text corpus ($\sim$2 minutes). The $\model$ involves two retrieval stages (T2I, I2T) for each $\sonetomany\onevar$ and $\fret\twovar$, each requiring approximately 0.4 hours. Calculating the final alignment scores with $\fret\twovar$ takes less than 2 minutes.

In total, refining the entire $\mathcal{D}_\text{SynthImgCap}$ for COCO with 542,401 image-caption pairs using $\model$ requires approximately 2.3 RTX A6000 GPU hours. $\model$ requires 1.4h more compared to CLIPScore while improving the performance with \textbf{7.6} CIDEr scores. Other web data pruning methods involve large models like Vicuna, LLaVA, or BLIP~\cite{vicuna, llava, li2022blip}; the preprocessing requires higher GPU time.

\subsection{$\model$ Application}

We also apply $\model$ in other zero-shot image captioning models to verify generalization. We modify text-only training image captioning models, CapDec, ViECap, and IFCap~\cite{capdec, viecap, ifcap} for training with synthetic image caption pairs. For CapDec and ViECap, we modify the CLIP text encoder into the CLIP image encoder and eliminate the noise injection procedure. For IFCap, we also perform the same method as CapDec and ViECap, replacing the Image-like Retrieval process of IFCap with standard image-to-text retrieval.

\noindent\textbf{$\model$ in Other Zero-shot Image Captioning Models.} We demonstrate the effectiveness of the $\model$ by comparing the performance between modified CapDec, ViECap, and IFCap~\cite{capdec, viecap, ifcap} trained on $\mathcal{D}_\text{SynthImgCap}$ and refined $\mathcal{D}_\text{SynthImgCap}^\star$ with $\model$. We report the MS-COCO in-domain performance in \Tref{tb:mm_zsic+ours}. As in previous experiments, $\model$ improves various zero-shot image captioning models, which validates the general applicability of $\model$. 

\subsection{Ablation Studies}

% We conduct comprehensive ablation studies to validate the design choices incorporated into $\model$. These studies systematically evaluate the impact of: 1) the core selection strategy $\s\onevar$ and scoring function $\f\twovar$, 2) the choice of Vision-Language Model (VLM) encoder used for cross-modal operations, 3) the text encoder $\mathcal{E}_T(\cdot)$ employed within the scorer $\fret\twovar$, 4) the preservation fraction $\tau$, and 5) the retrieval hyperparameters $K$ and $K_r$

\noindent\textbf{Selection Function.} We compare our primary one-to-many caption-to-image strategy ($\sonetomany\onevar$, detailed in~\Sref{method3.2}) against several alternatives: a baseline one-to-one mapping $\sonetoone\onevar$, and other one-to-many variants including text-to-text $S_\text{T2T}$, retrieve similar captions, then select their corresponding images, $S_\text{I2T}$ and $S_\text{I2I}$. Results presented in \Tref{tb:mm_ablationSelection} demonstrate that strategies using the caption as the primary query ($\sonetomany\onevar$ and $S_\text{T2T}$) significantly outperform those using the image as the query. This supports our hypothesis that focusing on high-quality captions as queries and potentially discarding poorly aligned synthetic images is more effective for refining image-caption datasets. 

\begin{table}[t]
\caption{Ablation studies of selection function $\s\onevar$ and multi-modal scorer function $\f\twovar$.}
\tabcolsep=8pt
\resizebox{0.8\columnwidth}{!}{
\begin{tabular}{c|c|ccccc}
\toprule[1.2pt]
\multirow{2}{*}{$\s$} & \multirow{2}{*}{$\f$} & \multicolumn{5}{c}{\textbf{COCO}} \\
 & & B@4 & M & R & C & S  \\
\midrule
\multicolumn{1}{l}{$\Baseline$ \cite{pcm-net}} &  & 31.5 & 25.9 & 53.9 & 103.8 & 19.7 \\
\midrule
\multirow{2}{*}{$\sonetoone$} & $\fsigliptwo$ & 31.7 & 25.6 & 53.6 & 104.0 & 19.4 \\
 & $\fret$ & 31.7 & 25.7 & 53.8 & 105.2 & 19.6 \\
\midrule
\multirow{2}{*}{$S_\text{T2T}$} & $\fsigliptwo$ & 32.3 & 26.0 & 54.2 & 106.8 & 20.0 \\
 & $\fret$ & 32.8 & 26.2 & 54.6 & 108.4 & 20.3 \\
\midrule
\multirow{2}{*}{$S_\text{T2I}$} & $\fsigliptwo$ & 32.9 & 26.2 & 54.7 & 108.2 & 20.3 \\
 & $\fret$ & \mff{33.6} & \mff{26.7} & \mff{55.3} & \mff{112.0} & \mff{20.5} \\
\midrule
\multirow{2}{*}{$S_\text{I2T}$} & $\fsigliptwo$ & 31.6 & 25.6 & 53.9 & 104.0 & 19.4 \\
 & $\fret$ & 31.5 & 25.7 & 53.9 & 104.8 & 19.7 \\
\midrule
\multirow{2}{*}{$S_\text{I2I}$} & $\fsigliptwo$ & 32.0 & 25.8 & 54.1 & 104.4 & 19.6 \\
 & $\fret$ & 31.8 & 25.7 & 53.9 & 105.2 & 19.6 \\
% \midrule
% \multicolumn{1}{l}{$\Baseline$ \cite{pcm-net}} & & 31.5 & 25.9 & 53.9 & 103.8 & 19.7 \\
% \multirow{2}{*}{$\sonetoone$}       & $\fsigliptwo$ &  31.7 & 25.6 & 53.6 & 104.0 & 19.4 \\
%                                     & $\fret$       &  31.7 & 25.7 & 53.8 & 105.2 & 19.6\\
% \midrule
% \multirow{2}{*}{$S_\text{T2T}$}     & $\fsigliptwo$ & 32.3 & 26.0 & 54.2 & 106.8 & 20.0 \\
%                                     & $\fret$  & 32.8 & 26.2 & 54.6 & 108.4 & 20.3 \\
% \midrule
% \multirow{2}{*}{$S_\text{T2I}$}     & $\fsigliptwo$ & 32.9 & 26.2 & 54.7 & 108.2 & 20.3 \\
%                                     & $\fret$  & 33.6 & 26.7 & 55.3 & 112.0 & 20.5 \\
% \midrule
% \multirow{2}{*}{$S_\text{I2T}$}     & $\fsigliptwo$ & 31.6 & 25.6 & 53.9 & 104.0 & 19.4 \\
%                                     & $\fret$  & 31.5 & 25.7 & 53.9 & 104.8 & 19.7 \\
% \midrule
% \multirow{2}{*}{$S_\text{I2I}$}     & $\fsigliptwo$ & 32.0 & 25.8 & 54.1 & 104.4 & 19.6 \\
%                                     & $\fret$  & 31.8 & 25.7 & 53.9 & 105.2 & 19.6 \\
\bottomrule[1.2pt]
\end{tabular}
}\label{tb:mm_ablationSelection}
\end{table}

\begin{figure}[t]
    \centering    
    \includegraphics[width=0.8\linewidth]{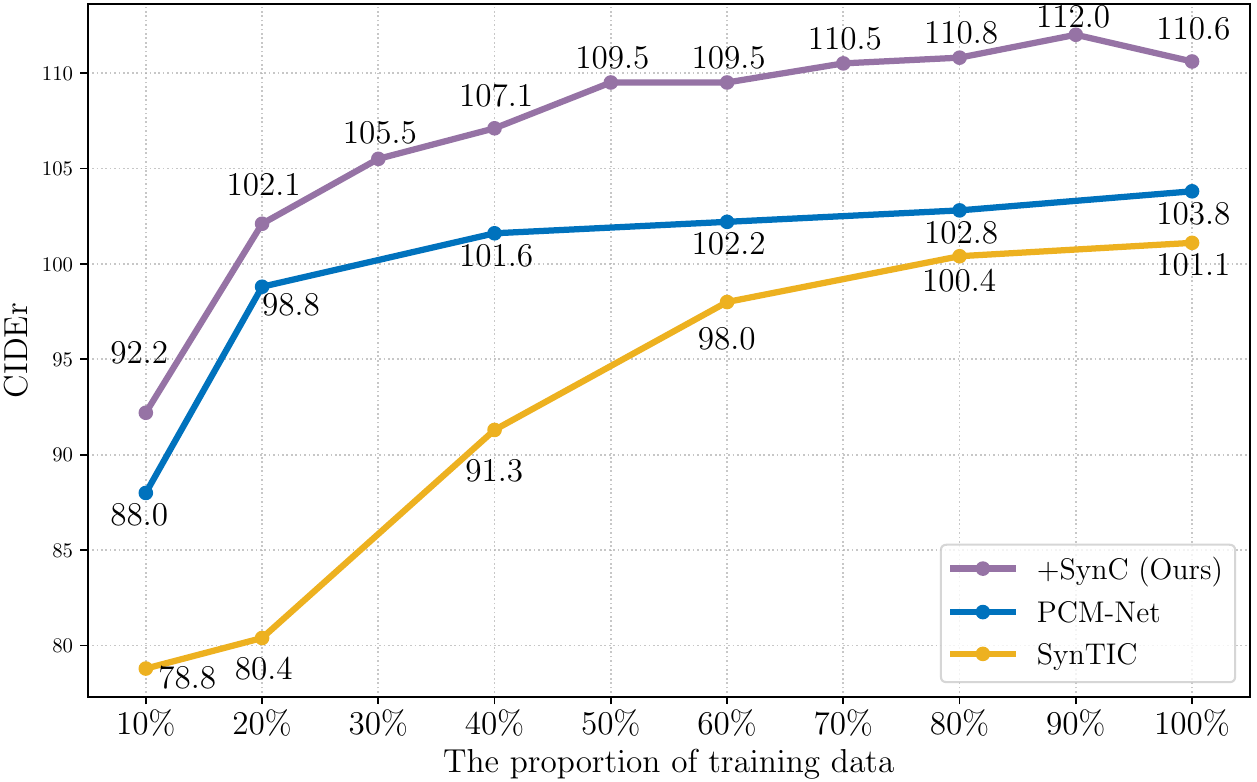}
    \caption{Ablation studies on preserving ratio $\tau$ on MS-COCO.}
    \label{fig:mm_reserve_ratio_pcmnet}
\end{figure}

\noindent\textbf{Multi-modal Scorer Function.} To isolate the effectiveness of the proposed $\fret\twovar$, we compare it against $\fsigliptwo\twovar$, a modified CLIPScore~\cite{clip} where the CLIP encoders are replaced with SigLIP2 ViT-B/16@256~\cite{siglip2} (matching the encoder used in our method for a fair comparison). This comparison is performed across the different selection strategies. As reported in \Tref{tb:mm_ablationSelection}, $\fret\twovar$ achieves superior performance across most metrics, especially when paired with the effective caption-query selection strategies, $\sonetomany\onevar$ and $S_\text{T2T}\onevar$. $\fret\twovar$ also performs comparably or slightly better than $\fsigliptwo\twovar$, even with less optimal selection methods. Based on these results, we confirm that the combination of the $\sonetomany\onevar$ selection strategy and the $\fret\twovar$ scoring function is the most effective configuration for $\model$.

\noindent\textbf{Choice of VLM Encoder for Cross-Modal Operations.}
We evaluate the influence of the VLM encoder used for the cross-modal retrieval in \Tref{tb:mm_ablationbackbone}. We compare 3 VLMs with various variants. While all tested encoders result in performance gains over the $\Baseline$, the SigLIP2 models yield the best results. Comparing the SigLIP2 variants, ViT-L/16 offers only marginal improvements over ViT-B/16 but incurs a substantially higher computational cost for embedding, approximately a 5 times increase. Considering this trade-off, we select SigLIP2 ViT-B/16@256 as the VLM encoder for $\model$ due to its strong performance and better efficiency.

\begin{table}[t]

\caption{Ablation studies on VLMs encoder $\mathcal{E}_I$ and $\mathcal{E}_T$.}
\tabcolsep=4pt
\resizebox{0.8\columnwidth}{!}{
\begin{tabular}{l|c|ccccc}
\toprule[1.2pt]
\multicolumn{2}{c|}{VLMs Encoder} & \multicolumn{5}{c}{\textbf{COCO}} \\
\multicolumn{1}{c}{Method} & Variant & B@4 & M & R & C & S \\
\midrule
\multicolumn{1}{l}{$\Baseline$~\cite{pcm-net}} & & 31.5 & 25.9 & 53.9 & 103.8 & 19.7 \\
\midrule
\multirow{3}{*}{CLIP~\cite{clip}} & ViT-B/32 & 31.9 & 25.7 & 54.2 & 105.9 & 19.6 \\
 & ViT-L/14 & 32.3 & 25.9 & 54.4 & 106.7 & 19.5 \\
 & RN50x4 & 32.0 & 25.8 & 54.2 & 106.8 & 19.6 \\
\cmidrule{1-2}
\multirow{2}{*}{SigLIP~\cite{siglip}} & ViT-B/16 & 33.3 & 26.6 & 55.1 & 109.9 & 20.6 \\
 & ViT-L/16@384 & 33.3 & 26.5 & 55.1 & 110.0 & 20.3 \\
\cmidrule{1-2}
\multirow{2}{*}{SigLIP2~\cite{siglip2}} & ViT-B/16@256 & 33.6 & 26.7 & 55.3 & \mff{112.0} & 20.5 \\
 & ViT-L/16@512 & \mff{34.0} & \mff{26.8} & \mff{55.5} & \mff{112.0} & \mff{20.8} \\
% $\Baseline$~\cite{pcm-net} &  & 31.5 & 25.9 & 53.9 & 103.8 & 19.7 \\
% \multirow{3}{*}{CLIP~\cite{clip}}       & ViT-B/32     & 31.9 & 25.7 & 54.2 & 105.9 & 19.6 \\
%                                         & ViT-L/14     & 32.3 & 25.9 & 54.4 & 106.7 & 19.5 \\
%                                         & RN50x4       & 32.0 & 25.8 & 54.2 & 106.8 & 19.6 \\
% \cmidrule{1-2}
% \multirow{2}{*}{SigLIP~\cite{siglip}}   & ViT-B/16     & 33.3 & 26.6 & 55.1 & 109.9 & 20.6 \\
%                                         & ViT-L/16@384 & 33.3 & 26.5 & 55.1 & 110.0 & 20.3 \\
% \cmidrule{1-2}
% \multirow{2}{*}{SigLIP2~\cite{siglip2}} & ViT-B/16@256 & 33.6 & 26.7 & 55.3 & 112.0 & 20.5 \\
%                                         & ViT-L/16@512 & 34.0 & 26.8 & 55.5 & 112.0 & 20.8 \\
\bottomrule[1.2pt]
\end{tabular}
}\label{tb:mm_ablationbackbone}
\end{table}
  
\begin{table}[t]

\caption{Ablation studies on text encoder $\mathcal{E}_S$ for $\fret\twovar$.}
\tabcolsep=8pt
\resizebox{0.8\columnwidth}{!}{
\begin{tabular}{l|ccccc}
\toprule[1.2pt]
\multirow{2}{*}{$\mathcal{E}_S$ for $\fret$} &  \multicolumn{5}{c}{\textbf{COCO}} \\

 & B@4 & M & R & C & S  \\
% & B@4 & M & C & S & B@4 & M & C & S \\
\midrule
$\Baseline$ \cite{pcm-net} & 31.5 & 25.9 & 53.9 & 103.8 & 19.7 \\
Sentence Transformer~\cite{wang2020minilm} & \mff{33.6} & \mff{26.7} & \mff{55.3} & \mff{112.0} & \mff{20.5} \\
SigLIP2 ViT-B/16@256~\cite{siglip2} & 33.1 & 26.4 & 54.9 & 108.6 & 20.3 \\
% $\Baseline$ \cite{pcm-net} & 31.5 & 25.9 & 53.9 & 103.8 & 19.7 \\
% Sentence Transformer~\cite{wang2020minilm}  &  33.6 & 26.7 & 55.3 & 112.0 & 20.5 \\
% SigLIP2 ViT-B/16@256~\cite{siglip2} & 33.1 & 26.4 & 54.9 & 108.6 & 20.3 \\
\bottomrule[1.2pt]
\end{tabular}
}\label{tb:mm_ablation_textencoder}
\end{table}

\begin{figure}[t]
    \centering    
    \includegraphics[width=0.9\linewidth]{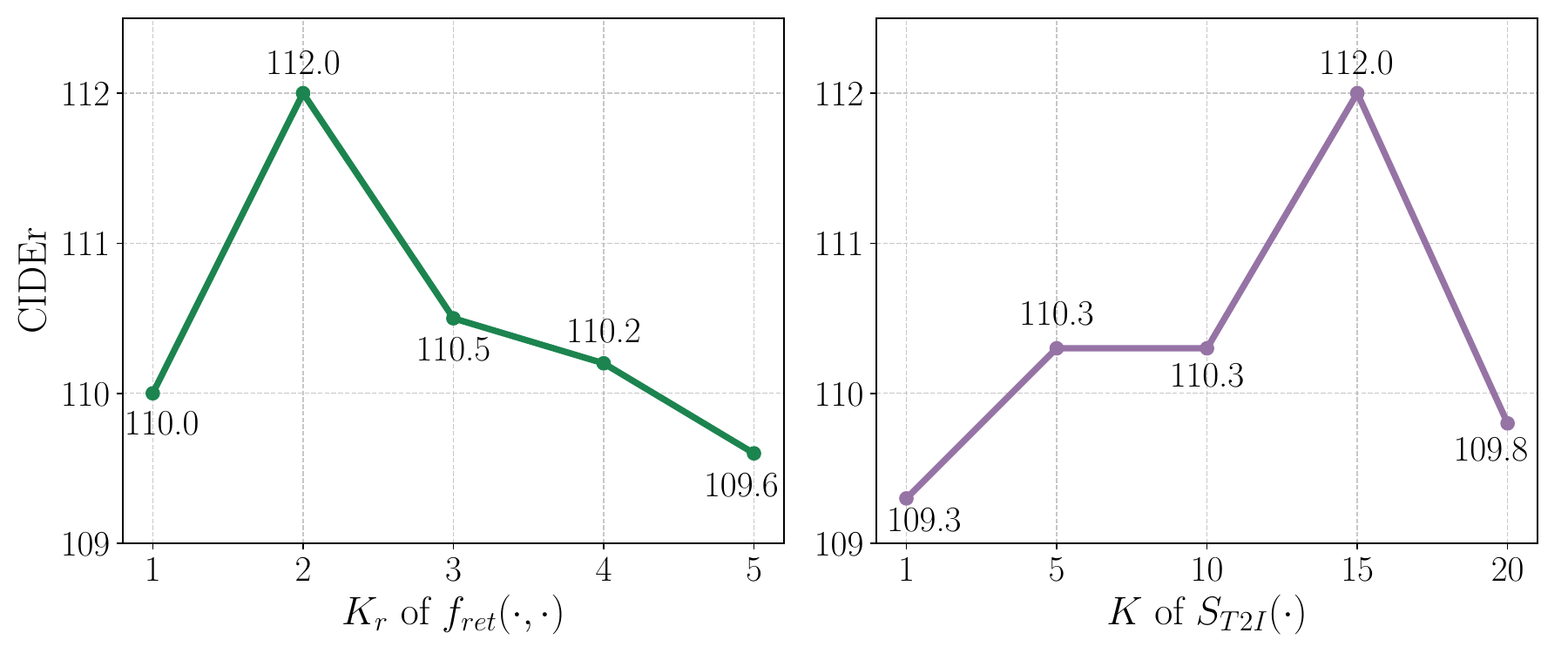}\caption{Ablation studies on hyperparameters $K$ and $K_r$.}
    \label{fig:mm_ablation}
\end{figure}

\noindent\textbf{Text encoder for $\fret\twovar$.} 
We specifically ablate the choice of the text encoder $\mathcal{E}_T(\cdot)$ for generating these embeddings within $\fret\twovar$ in \Tref{tb:mm_ablation_textencoder}. We compare using the text encoder from our chosen VLM (SigLIP2 ViT-B/16)~\cite{siglip2} and an unimodal text encoder, Sentence Transformer~\cite{wang2020minilm}. In line with findings from prior work~\cite{sieve} suggesting advantages for unimodal encoders in semantic similarity tasks, employing the Sentence Transformer within $\fret\twovar$ resulted in higher scores across all metrics.

\noindent\textbf{Hyperparameter, $\tau$, $K$, and $K_r$.} 
Finally, we examine the sensitivity of $\model$ to key hyperparameters. For preserving ratio $\tau$, we test on the COCO test split by increasing every 10\% from 10\% to 100\% of the training data. The result is in ~\Fref{fig:mm_reserve_ratio_pcmnet}, presenting that using only 30\% of the training dataset dominates the performance of the $\Baseline$. The best performance is when $\tau=0.9$. We evaluate various settings for $K$ and $K_r$, and the results are in ~\Fref{fig:mm_ablation}, indicating that $\model$ performs robustly across different settings. The configuration $K_r=2$ and $K=15$ shows the best performance overall.

\section{Conclusion}
This work tackles the critical challenge of semantic misalignment in T2I-generated synthetic datasets, which limits zero-shot image captioning (ZIC) performance. We introduce $\model$, a novel framework employing a one-to-many mapping strategy $\sonetomany\onevar$ and the multi-modal scorer function $\fret\twovar$ to refine noisy synthetic image-caption paired datasets. Our extensive experiments validate $\model$, demonstrating substantial and consistent performance gains across diverse ZIC models and benchmarks, including state-of-the-art results. $\model$ provides a practical approach to improve the quality and utility of synthetic data for ZIC. Its principles also hold promise for broader application, and future work will investigate extending $\model$ to other vision-and-language downstream tasks (e.g., segmentation~\cite{Kim2025SIDA}, visual question answering~\cite{cho2023generative}) and employ advanced image generative models~\cite{Oh2025catch, cha2025verbdiff}.

\begin{acks}
  This was partly supported by the Institute of Information \& Communications Technology Planning \& Evaluation (IITP) grant funded by the Korean government(MSIT) (No.RS-2020-II201373, Artificial Intelligence Graduate School Program(Hanyang University)) and the Institute of Information \& Communications Technology Planning \& Evaluation (IITP) grant funded by the Korean government(MSIT) (No.RS-2025-02219062, Self-training framework for VLM-based defect detection and explanation model in manufacturing process).
\end{acks}

\bibliographystyle{ACM-Reference-Format}
\bibliography{sample-base}

% \appendix

% \input{sections/X_suppl}

\end{document}